# Learning Sparse Causal Models is not NP-hard

Tom Claassen, Joris M. Mooij, and Tom Heskes
Institute for Computer and Information Science
Radboud University Nijmegen
The Netherlands

## Abstract

This paper shows that causal model discovery is *not* an NP-hard problem, in the sense that for sparse graphs bounded by node degree $k$ the sound and complete causal model can be obtained in worst case order $N^{2(k+2)}$ independence tests, even when latent variables and selection bias may be present. We present a modification of the well-known FCI algorithm that implements the method for an independence oracle, and suggest improvements for sample/real-world data versions. It does not contradict any known hardness results, and does not solve an NP-hard problem: it just proves that sparse causal discovery is perhaps more complicated, but not as *hard* as learning minimal Bayesian networks.

## 1 Introduction

Causal discovery is one of the cornerstones behind scientific progress. In recent years, significant breakthroughs have been made in causal inference under very reasonable assumptions, even when only data from observations are available (Pearl, 2000; Spirtes et al., 2000). Still, it is probably safe to say that many researchers consider causal discovery to be a difficult problem, and that it is generally thought to be computationally at least as hard as related problems such as learning minimal Bayesian networks.

The class of NP problems (non-deterministic, polynomial time) are problems for which a solution can be *verified* in polynomial time: order $O(N^k)$ for some constant $k$ given input size $N$, but for which no known polynomial time algorithm exists (or is thought to exist) that *finds* such a solution. In many cases algorithms exist that are able to find a solution quickly, or at least provide good approximations, but still have worst-case exponential running time order $O(2^{N^k})$.

Typical examples include finding the shortest path through all nodes in a weighted graph (travelling salesman), coloring vertices in a graph so that no two adjacent vertices share the same color, probabilistic inference in a Bayesian network (Cooper, 1990), and the Boolean satisfiability problem ($k$-SAT, the first known NP-complete problem).

A problem is NP-hard if it is at least as hard as the hardest problems in NP. Put differently, a problem is NP-hard if there is a polynomial time reduction of an NP-complete problem to it, so that any polynomial time solution to the NP-hard problem implies that all NP-complete problems can be solved in polynomial time. See (Garey and Johnson, 1979) for a standard introduction to the subject, and e.g. (Goldreich, 2008) for a more modern approach.

In this article we focus on the problem of learning a causal model from probabilistic information. We assume there is a system that is characterized by a so-called causal DAG $\mathcal{G}_C$ that describes the causal interactions between the variables in the system. The structure of this causal DAG is invariant and responsible for a probability distribution over the subset of observed variables. The goal is to learn as much as possible about the presence or absence of certain causal relations between variables in the underlying causal DAG from the available probabilistic information.

Chickering et al. (2004) showed that finding an inclusion-optimal (minimal) Bayesian network for a given probability distribution is NP-hard, even when a constant-time independence oracle is available (see §2 for more details). Such hardness results have inspired many creative approaches to network learning, e.g. methods that seek to find efficient approximations to minimal Bayesian networks through greedy search (Chickering, 2002), or methods that employ specialized heuristics or solver techniques to make exact learning feasible from 30, up to even 60 variables if the graph is sufficiently sparse (Yuan and Malone, 2012; Cussens, 2011).

On the face of it, minimal Bayesian network inference seems close to a simplified and idealized version of a causal model discovery problem. Hence it may be unsurprising to find that currently available methods also have worst-case exponential complexity (see §3.2).

However, in this paper we show that causal model discovery in sparse graphs is in fact *not* an NP-hard problem, even in the presence of latent confounders and/or selection bias. We do this by providing an adaptation of the well-known FCI algorithm (see §3.2) for learning the sound and complete causal model in the form of a PAG with running time in the worst case polynomial in the number of independence tests to order $O(N^{2(k+2)})$, where $N$ is the number of variables and $k$ the maximum node degree in the causal model over the observed variables.

**Graphical causal models**

A *mixed graph* $\mathcal{G}$ is a graphical model that can contain three types of edges between pairs of nodes: directed ($\rightarrow$), bi-directed ($\leftrightarrow$), and undirected ($-$). In a mixed graph, standard graph-theoretical notions, e.g. *child/parent, ancestor/descendant, directed path, cycle*, still apply, with natural extension to sets. In particular $Adj_{\mathcal{G}}(\mathbf{X})$ indicates all nodes adjacent to (but not in) the set of nodes $\mathbf{X}$ in the graph $\mathcal{G}$, and $An_{\mathcal{G}}(\mathbf{X})$ indicates all (ancestors of) nodes in $\mathbf{X}$ in the graph $\mathcal{G}$. A vertex $Z$ is a *collider* on a path $u = \langle \ldots, X, Z, Y, \ldots \rangle$ if there are arrowheads at $Z$ on both edges from $X$ and $Y$, otherwise it is a *noncollider*.

A mixed graph $\mathcal{G}$ is *ancestral* iff an arrowhead at $X$ on an edge to $Y$ implies there is no directed path from $X$ to $Y$ in $\mathcal{G}$, and there are no arrowheads at nodes with undirected edges. As a result, arrowhead marks can be read as 'is not an ancestor of'. In a mixed graph $\mathcal{G}$, a vertex $X$ is *m-connected* to $Y$ by a path $u$, relative to a set of vertices $\mathbf{Z}$, iff every noncollider on $u$ is not in $\mathbf{Z}$, and every collider on $u$ is an ancestor of $\mathbf{Z}$. If there is no such path, then $X$ and $Y$ are *m-separated* by $\mathbf{Z}$. An ancestral graph is *maximal* (MAG) if for any two nonadjacent vertices there is a set that separates them. A *directed acyclic graph* (DAG) is a special kind of MAG, containing only $\rightarrow$ edges, for which $m$-separation reduces to the familiar $d$-separation criterion. The *Markov property* links the structure of an ancestral graph $\mathcal{G}$ to observed probabilistic independencies: $X \perp\!\!\!\perp Y \mid \mathbf{Z}$, if $X$ and $Y$ are $m$-separated by $\mathbf{Z}$. *Faithfulness* implies that the only observed independencies in a system are those entailed by the Markov property. For more details, see (Koller and Friedman, 2009; Spirtes et al., 2000).

A *causal DAG* $\mathcal{G}_C$ is a directed acyclic model where the arcs represent direct causal interactions (Pearl, 2000). In general, the independence relations between observed variables in a causal DAG can be represented in the form of a MAG (Richardson and Spirtes, 2002). The (complete) partial ancestral graph (PAG) represents all invariant features that characterize the equivalence class $[\mathcal{G}]$ of such a MAG, with a tail '$-$' or arrowhead '$>$' end mark on an edge, iff it is invariant in $[\mathcal{G}]$, otherwise it has a circle mark '$\circ$', see (Zhang, 2008). Figure 1 illustrates the relation between these three types of graphs.

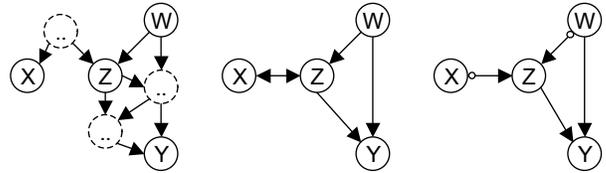

Figure 1: (a) Assumed underlying causal DAG $\mathcal{G}_C$; (b) corresponding MAG $\mathcal{M}$ over observed variables; (c) causal model as PAG, with $N = 4$ and $k \leq 3$.

**C-LEARN**

The completed PAG represents all valid causal information that can be inferred from independencies between observed variables: this will be the target causal model of the learning task (C-LEARN).

To distinguish between the contribution of the learning task and the calculation of the independencies themselves, we introduce the following notion from (Chickering et al., 2004):

**Definition 1.** An **independence oracle** for a distribution $p(\mathbf{X})$ is an oracle that, in constant time, can determine whether or not $X \perp\!\!\!\perp Y \mid \mathbf{Z}$, for any $(\{X,Y\} \cup \mathbf{Z}) \subseteq \mathbf{X}$.

The **C-LEARN** task can now be described as:
INSTANCE: Given an independence oracle $\mathcal{O}$ for a set of variables $\mathbf{X} = \{X_1, .., X_N\}$ that is faithful to an underlying causal DAG $\mathcal{G}_C$, and constant bound $k$.
TASK: Find the completed PAG model that matches $\mathcal{O}$ with node degree $\leq k$.

## 2 Why learning minimal Bayesian networks is NP-hard

It has long been known that learning a minimal Bayesian network is an NP-hard problem, even when an independence oracle is available and each node in the network has at most $k \geq 3$ parents. An elegant proof is provided in (Chickering et al., 2004) by introducing a polynomial reduction from an established NP-complete problem known as *Degree-Bounded Feedback Arc Set* (DBFAS) to an instance of learning a minimal Bayesian network (B-LEARN) with maximum node degree $k$. Given an arbitrary network of directed edges the target in DBFAS is to find the smallest set of arcs (the 'feedback set') whose removal eliminates all directed cycles from the graph.

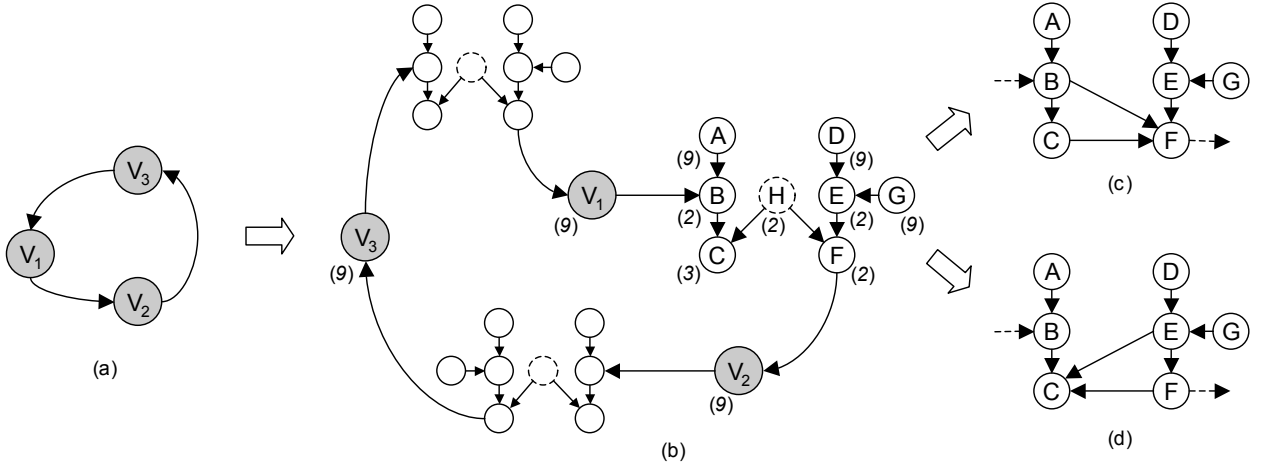

Figure 2: (a) DBFAS instance over 3 nodes; (b) corresponding Bayesian network reduction with distinct variables $\{A, B, C, D, E, F, G, H\}$ for each edge-component (nr. of states per variable in brackets below); (c) BN edge-component configuration with 16 parameters to specify the distribution over $C$ and $F$; (d) idem, with 18 parameters.

The reduction strategy from DBFAS to B-LEARN is depicted in Figure 2. It replaces the DBFAS instance with an equivalent Bayesian network, in which each directed edge $V_i \to V_j$ in DBFAS is replaced by an edge-component over discrete variables $\{A, B, C, D, E, F, G, H\}$. All DBFAS nodes $V_i$ and all nodes $\{A, D, G\}$ in each edge-component get 9 states, the nodes $\{B, E, F, H\}$ get 2 states, and only the nodes $C$ get 3 states. Furthermore, the $H$ nodes in each edge-component are supposedly hidden, i.e. not in the set of observed variables in the minimal Bayesian network from B-LEARN, see Figure 2.

The connection between the instance of DBFAS and B-LEARN relies on the fact that it is *impossible* to construct a Bayesian network that can faithfully represent the independence relations from the underlying causal DAG when the $H$ variables are not observed. As a result, each edge component is forced to choose between configuration (c) or (d) in Figure 2, with a preference for the smaller (c), *unless* this introduces a directed cycle due to the connection via $V_i \to B \to F \to V_j$. In that case the cycle can be broken by opting for (d). The (acyclic) minimal global Bayesian network will have the fewest edge-components oriented as (d) needed to break all cycles, and so all these edge components together represent a *minimal feedback arc set* for the original DBFAS problem.

The transformation from DBFAS to B-LEARN is polynomial, which implies that if there is any method that can solve B-LEARN in polynomial time, then it can also solve DBFAS in polynomial time, and with it the entire class of NP-complete problems. In particular, it also applies when there is a constant time independence oracle for B-LEARN and the max. node degree in the optimal solution is bounded by any $k \geq 3$. It is exactly this subclass of problems that are the focus of this paper in the context of learning a causal model.

### Causal model reduction

An obvious question would be why the previous result should not apply to causal models in general, especially given the importance, explicit or implicit, of minimality in many structure learning tasks.

The answer to this lies in the fact that the 'Bayesian network' requirement itself imposes the additional restriction on the solution that only directed edges are allowed, which does not come into play when causal models are concerned. Therefore B-LEARN cannot opt for the configuration depicted in Figure 3, due to the edge $C \leftrightarrow F$, whereas C-LEARN has no such problem. This structure is actually *smaller* than the corre-sponding minimal Bayesian network component, as it requires only 12 free parameters to specify the contribution of each $C-F$ pair in the network: $2 \times 2 \times 2 = 8$ for the $C$ node, $2 \times 2 \times 1 = 4$ for $F$, and 1 for the $H$ node, giving a total of 13 parameters, one of which can be marginalized out, see, e.g. (Evans and Richardson, 2010) for details on parameterizing acyclic directed mixed graphs (ADMGs), allowing for both directed and bi-directed edges.

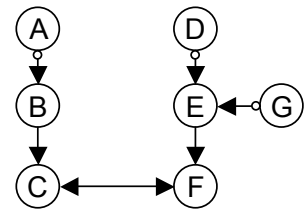

Figure 3: Minimal edge-component configuration in causal model, requiring 12 free parameters for variables $C$ and $F$.

As a result, a DBFAS reduction to C-LEARN along the lines of (Chickering et al., 2004) will fail to solve the original NP-complete problem, as the causal model will have *all* edge components oriented as in Figure 3, giving no information whatsoever on a minimal feedback arc set for the corresponding DBFAS instance.

We now turn to existing methods for causal discovery to see where they run into exponential trouble.

## 3 Independence based Causal Discovery

### 3.1 Causally sufficient systems

Many methods have been developed that efficiently learn the correct causal model when there are no unobserved common causes between the variables, e.g. IC (Pearl and Verma, 1991), PC (Spirtes et al., 2000), Grow-Shrink (Margaritis and Thrun, 1999), etc.

As a typical example, a version of PC is depicted in Algorithm 1. From a fully connected undirected graph $\mathcal{G}$, it consists of two stages: an adjacency search to remove edges, followed by an orientation phase. In the first stage, for each pair of nodes $X - Y$ (still) connected in $\mathcal{G}$ it searches for a subset of adjacent nodes $\mathbf{Z}$ that can separate them: $X \perp\!\!\!\perp Y \,|\, \mathbf{Z}$; if found the edge is removed. By checking all adjacent node-pairs in $\mathcal{G}$ for possible separating sets of increasing size, the PC algorithm ensures that it finds separating sets as small as possible. If the node degree in the true causal model $\mathcal{G}$ is bounded by $k$, then worst case it needs to check all subsets size $1..k$ from $N$ nodes, for $N^2$ nodes on edges, resulting in a polynomial running time dominated by order $O(N^{k+2})$ for the adjacency search. Afterwards an orientation phase adds all invariant edge-marks (tails or arrowheads) by rules that trigger on the existence of certain path-configurations in $\mathcal{G}$, which can be checked in order $O(N^3)$ (Spirtes et al., 2000).

### 3.2 Causal discovery with latent confounders

Unfortunately, the PC algorithm can run into trouble when applied to causal models where causal sufficiency is not guaranteed. In that case it can miss certain separating sets that may require nodes not adjacent to either of the separated nodes. Figure 4(b) depicts the canonical 5-node example, where the edge $X - Y$ can be eliminated by the set $\{U, V, Z\}$, but this is not found by the PC algorithm, as at that stage $Z$ is no longer adjacent to $X$ or $Y$. As a result, edges may fail to be eliminated from $\mathcal{G}$, possibly leading to erroneous causal conclusions in the orientation stage, see e.g. (Colombo et al., 2012, §3) for more examples.

**Algorithm 1** PC-algorithm
  **In** : independence oracle $\mathcal{O}$ for variables $\mathbf{V}$
  **Out**: causal model $\mathcal{G}$ over $\mathbf{V}$
  *Adjacency search*
1: $\mathcal{G} \leftarrow$ fully connected undirected graph over $\mathbf{V}$
2: $n = 0$
3: **repeat**
4:   **repeat**
5:     select $X$ with $|Adj_\mathcal{G}(X)| > n$,
6:     select $Y \in Adj_\mathcal{G}(X)$
7:     **repeat**
8:       select subset $\mathbf{Z}$ size $n$ from $Adj_\mathcal{G}(X) \setminus Y$,
9:       **if** $X \perp\!\!\!\perp Y \,|\, \mathbf{Z}$ **then**
10:         $Sepset(X,Y) = Sepset(Y,X) = \mathbf{Z}$
11:         remove edge $X - Y$ from $\mathcal{G}$
12:       **end if**
13:     **until** all subsets size $n$ have been tested
14:   **until** all edges $X - Y$ in $\mathcal{G}$ have been checked
15:   $n = n + 1$
16: **until** no more nodes with $|Adj_\mathcal{G}(X)| > n$
  *Orientation phase*
17: **for all** unshielded triples $X - Z - Y$ in $\mathcal{G}$ **do**
18:   **if** $Z \notin Sepset(X,Y)$ **then**
19:     orient $v$-structure $X \rightarrow Z \leftarrow Y$
20: **end for**
21: run other orientation rules until no more new
22: **return** causal model $\mathcal{G}$

To tackle this problem, Spirtes et al. (1999) developed the so-called Fast Causal Inference (FCI) algorithm that introduces an additional stage to the adjacency search. It searches for an extended set of nodes: the **D-SEP**$(A, B)$ set, roughly corresponding to ancestors of $\{A, B\}$ that are adjacent to $A$ and/or reachable via a bi-directed path, for which it can be shown that (with observed variables $\mathbf{O}$ and selection set $\mathbf{S}$):

**Lemma 1.** (Spirtes et al., 1999, Lemma 12) *If there is some subset $\mathbf{W} \subseteq \mathbf{O} \setminus \{A, B\}$ such that $A$ and $B$ are $d$-separated by $\mathbf{W} \cup \mathbf{S}$, then $A$ and $B$ are $d$-separated given $\mathbf{D\text{-}SEP}(A, B) \cup \mathbf{S}$.*

The problem is that at that stage it is not yet known which nodes exactly belong to $An(\{A, B\})$. The solution employed by FCI is as follows: after the initial PC-adjacency search it adds some orientation informa-tion to the resulting adjacency graph to obtain a partially oriented graph $\pi_0$, and from this identifies the set **Possible-D-SEP**$(A, B, \pi_0)$ for edge $A - B$ that is a guaranteed superset of **D-SEP**$(A, B)$. It then tests for all subsets of **Possible-D-SEP**$(A, B, \pi_0)$ until a separating set is found, or not, in which case the edge should remain present.

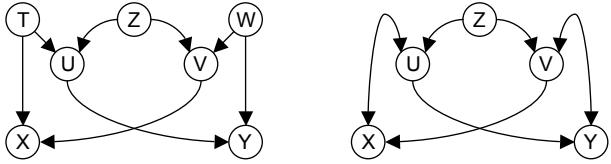

Figure 4: (a) With causal sufficiency (no confounders) we only need to check subsets from $Adj(X)$ or $Adj(Y)$, e.g. $X \perp\!\!\!\perp Y \,|\, V, T$; (b) Without causal sufficiency node $Z \notin Adj(X,Y)$ is needed in the separating set: $X \perp\!\!\!\perp Y \,|\, U, V, Z$

But even for graphs with degree $\leq k$, neither the size of the **Possible-D-SEP**$(A, B, \pi_0)$ set, nor the size of the target **D-SEP** set is bounded by $k$. As a result, the worst case is now dominated by a term that could involve searching through all subsets from $N$ nodes, which brings it back to exponential order time complexity $O(2^N)$.

In practice, the number of edges removed in the FCI stage tends to be very small (relative to the PC-stage), even though it is often the most expensive part of the algorithm by far. This behaviour is exploited effectively in the RFCI algorithm (Colombo et al., 2012): it skips the additional FCI stage and ensures the output is still sound (though no longer necessarily complete) through a modified orientation phase that avoids some of the erroneous causal conclusions PC could make when edges are missed.

In this article we take a different approach by showing that it is possible to build up an alternative **D-SEP** set that is completely determined by nodes adjacent to $\{A, B\}$ and already inferred minimal separating sets from the PC-adjacency search.

## 4 D-separating sets

First some terminology for the target nodes and edges:

**Definition 2.** In a MAG $\mathcal{M}$, two nodes $X$ and $Y$ are **D-separated** by $\mathbf{Z}$ iff they are $d$-separated by $\mathbf{Z}$, and all sets that *can* separate $X$ and $Y$ contain at least one node $Z \notin Adj(\{X, Y\})$. Such a node $Z \in \mathbf{Z}$ that cannot be made redundant by nodes adjacent to $X$ or $Y$ is a **D-sep node**, and $(X, Y)$ is a **D-sep link**,

For example in Figure 4(b), $X$ and $Y$ are $D$-separated by $\{U, V, Z\}$, including $D$-sep node $Z$.

After the PC-adjacency search all $D$-sep links are still connected by an edge in the skeleton $\mathcal{G}$. Below we first discuss how to recognize possible $D$-sep links, and then how to find an appropriate $D$-separating set, including the elusive $D$-sep nodes. The third part puts it all together into a sound and complete search strategy. Proof details can be found in the Appendix and in (Claassen et al., 2013).

### 4.1 Identifying D-sep links

To characterize $D$-separable nodes, we use a result from (Spirtes et al., 1999; Claassen and Heskes, 2011):

**Lemma 2.** For disjoint (subsets of) nodes $X, Y, Z, \mathbf{Z}$ from the observed variables $\mathbf{O}$ in a causal graph $\mathcal{G}$ with selection set $\mathbf{S}$,

(1) $X \not\perp\!\!\!\perp Y \,|\, \mathbf{Z} \cup [Z] \;\;\Rightarrow\;\; Z \notin An_{\mathcal{G}}(\{X, Y\} \cup \mathbf{Z} \cup \mathbf{S})$.

(2) $X \perp\!\!\!\perp Y \,|\, [\mathbf{Z} \cup Z] \;\;\Rightarrow\;\; Z \in An_{\mathcal{G}}(\{X, Y\} \cup \mathbf{S})$,

where square brackets indicate a *minimal* set of nodes.

Rule (1) identifies invariant arrowheads on edges; rule (2) is used in §4.2 to build up a $D$-separating set. With Lemma 2 it is easy to show the following properties:

**Lemma 3.** In a MAG $\mathcal{M}$, if two nodes $X$ and $Y$ are $D$-separated by a minimal set $\mathbf{Z}$, then:

1. $X \notin An(\{Y\} \cup \mathbf{Z} \cup \mathbf{S})$
2. $Y \notin An(\{X\} \cup \mathbf{Z} \cup \mathbf{S})$
3. $\forall Z \in \mathbf{Z} : Z \in An(\{X, Y\} \cup \mathbf{S})$

It motivates the following introduction:

**Definition 3.** The **Augmented Skeleton** $\mathcal{G}^+$ is obtained from the skeleton $\mathcal{G}$ by adding all invariant arrowheads that follow from single node minimal dependencies $X \not\perp\!\!\!\perp Y \,|\, \mathbf{Z} \cup [W]$ by Lemma 2, rule(1).

By testing for dependence on adding single nodes to the minimal separating sets from the PC-adjacency search all invariant arrowheads in $\mathcal{G}^+$ are found. Note that only nodes $Z$ adjacent to $\{X, Y\} \cup \mathbf{Z}$ in $\mathcal{G}$ need to be tested, and that all arrowheads from FCI's partially oriented graph $\pi_0$ (see §3.2) are also in $\mathcal{G}^+$.

The important implication is that $D$-sep links take on a very distinct pattern in $\mathcal{G}^+$:

**Lemma 4.** For a MAG $\mathcal{M}$, let $X$ and $Y$ be $D$-separable nodes that are adjacent in the corresponding augmented skeleton $\mathcal{G}^+$. If there are no edges in $\mathcal{G}^+$ between (other) $D$-separable nodes in $An(\{X, Y\})$, then $\mathcal{G}^+$ contains the following pattern: $U \leftrightarrow X \leftrightarrow Y \leftrightarrow V$, with $U$ and $V$ not adjacent in $\mathcal{G}^+$, and paths $V .. \rightarrow X$ and $U .. \rightarrow Y$ that do not go against an arrowhead.

For example, in the augmented skeleton in Figure 5(b), $D$-sep link $X - Z$ occurs in pattern $S \leftrightarrow X \leftrightarrow Z \leftrightarrow T$.

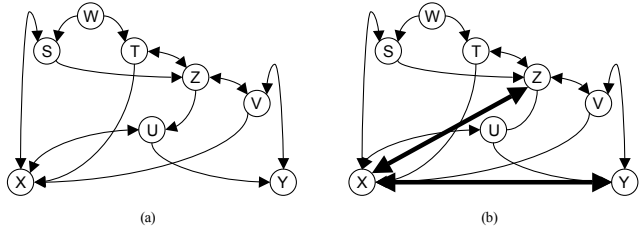

Figure 5: (a) Hierarchical $D$-sep links: $X \perp\!\!\!\perp Z \,|\, [S,T,W]$, with $Z$ also appearing in $X \perp\!\!\!\perp Y \,|\, [S,T,U,V,W,Z]$; (b) corresponding augmented skeleton $\mathcal{G}^+$, with initially undiscovered $D$-sep links $X \leftrightarrow Z$ and $X \leftrightarrow Y$ (bold)

There is a hierarchical structure between $D$-sep links in the sense that if one pair of $D$-separated nodes is part of the $D$-separating set of another, then not the other way around.

**Lemma 5.** In a MAG $\mathcal{M}$, for two pairs of $D$-separable nodes $X \perp\!\!\!\perp Y \,|\, [\mathbf{Z}]$ and $X \perp\!\!\!\perp Y \,|\, [\mathbf{W}]$, if $X \in \mathbf{W}$ and/or $Y \in \mathbf{W}$, then $U \notin \mathbf{Z}$ and/or $V \notin \mathbf{Z}$.

It implies that we may need to find one $D$-sep link in a MAG $\mathcal{M}$ before we can find another, but there is no vicious circle of complex intertwined $D$-sep links.

Though all this will facilitate (and speed up) identification, it is not enough for a polynomial search algorithm, as this is dominated by the number of possible node-sets to consider for a single $D$-sep link.

### 4.2 Capturing the $D$-sep nodes

In this part we first show that all $D$-sep nodes needed to separate $(X,Y)$ were already found as part of a minimal separating set between ancestors of $\{X,Y\}$. Then we show that we can recursively include these to obtain a $D$-separating set, starting from an appropriate set of nodes adjacent to $X$ and $Y$.

Using shorthand $AA(\mathbf{X}) \equiv (Adj(\mathbf{X}) \cap An(\mathbf{X})) \setminus \mathbf{X}$ for the set of **adjacent ancestors** of $\mathbf{X}$, we find that:

**Lemma 6.** In a MAG $\mathcal{M}$, if $Z \in \mathbf{Z}$ is a $D$-sep node in $X \perp\!\!\!\perp Y \,|\, [\mathbf{Z}]$, then $Z$ is also part of a minimal separating set between another pair of nodes from $\{X,Y\} \cup \mathbf{Z}_{\setminus Z} \cup AA(\{X,Y\})$, neither of which have selection bias.

As a result, if we know all minimal conditional independencies between nodes in $An(\{X,Y\})$, then all required $D$-separating nodes for $X \perp\!\!\!\perp Y \,|\, [\mathbf{Z}]$ already appear in one of these minimal separating sets. We want to use this information to guide the search for $D$-separating sets, but unfortunately we do not know what the ancestors are, and we do not have all minimal independencies (as the PC search only finds one minimal separating set for each eliminated edge).

Fortunately we can show that we do not need to know all minimal separating sets to find the relevant nodes, as it is sufficient to have just *one* minimal separating set for each nonadjacent pair, and recursively include these in the set. In formal terms:

**Definition 4.** An **independence set** $\mathcal{I} \subseteq \mathcal{I}(\mathcal{M})$ is a (sub)set of all minimal independence statements consistent with MAG $\mathcal{M}$, which contains at least one separating set for each pair of nonadjacent nodes in $\mathcal{M}$.

And a recursive definition for a set of separating nodes:

**Definition 5.** Let $\mathcal{I}$ be an independence set, then for a set $\mathbf{X}$ the **hierarchy** $HIE(\mathbf{X},\mathcal{I})$ is the union of $\mathbf{X}$ and all nodes that appear in a minimal separating set in $\mathcal{I}$ between any pair of nodes in $HIE(\mathbf{X},\mathcal{I})$.

By Lemma 2 all nodes in $HIE(\mathbf{X},\mathcal{I})$ are ancestors of one or more nodes in $\mathbf{X}$; see also Example 1, below.

With this the key result can be stated as:

**Lemma 7.** Let $X$ and $Y$ be $D$-separable nodes in a MAG $\mathcal{M}$. If independence set $\mathcal{I}$ contains at least one minimal separating set for each pair of nonadjacent nodes in $An(\{X,Y\})$ in $\mathcal{M}$ (except for $\{X,Y\}$ itself), then $HIE\bigl(AA(\{X,Y\}),\mathcal{I}\bigr)_{\setminus\{X,Y\}}$ is a $D$-separating set for $X$ and $Y$.

*In words*: if $X$ and $Y$ are $D$-separable nodes, then they are separated by the hierarchy of minimal separating nodes implied by ancestors adjacent to $X$ and/or $Y$.

If found, then the edge $X \leftrightarrow Y$ is removed from $\mathcal{G}^+$, and a corresponding minimal separating set is obtained by eliminating redundant nodes one by one until no more can be removed (Tian et al., 1998).

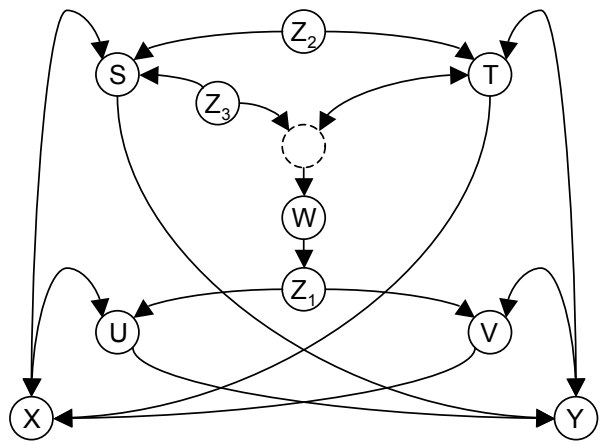

Figure 6: Hierarchical inclusion of separating sets ensures inclusion of $D$-sep node $Z_3$.

**Example 1.** In Figure 6, $X$ and $Y$ are $D$-separated by the set $\{S,T,U,V,Z_1,Z_2,Z_3\}$. Lemma 6 states that we can find $D$-sep node $Z_3$ from $S \perp\!\!\!\perp Z_1 \,|\, [Z_3]$, but the PC-stage may have found $S \perp\!\!\!\perp Z_1 \,|\, [W]$ instead, which would leave path $X \leftrightarrow S \leftrightarrow W \leftrightarrow T \leftrightarrow Y$ unblocked. However, Lemma 7 ensures that, whatever the independencies found by PC, node $Z_3$ is included in $HIE(\{X,Y,S,T,U,V\},\mathcal{I})$; indeed, we find $Z_3$ from either $S \perp\!\!\!\perp Z_1 \,|\, [Z_3]$ and/or $S \perp\!\!\!\perp W \,|\, [Z_3]$.

At this point, for a given suspect $D$-sep link $X \leftrightarrow Y$ in $\mathcal{G}^+$ we do not know exactly which nodes are in $AA(\{X,Y\})$ in $\mathcal{M}$. However, given that $\forall\{X,Y,Z\}$:

1. $AA(\{X,Y\}) \subseteq Adj(X) \cup Adj(Y)$,
2. $|Adj(X)| \leq k$ in $\mathcal{M}$, and
3. $Z \in Adj(X)$ in $\mathcal{M}$ $\Rightarrow$ $Z \in Adj(X)$ in $\mathcal{G}^+$,

we can be sure that the set $AA(\{X,Y\})$ in $\mathcal{M}$ consists of a combination of *at most* $k$ nodes from $Adj(X)$ and $k$ nodes from $Adj(Y)$ in $\mathcal{G}^+$.

For bounded $k$ those can be searched in worst case polynomial time $N^k \times N^k = N^{2k}$, and so, if $X \leftrightarrow Y$ is indeed a $D$-sep link, we are guaranteed to find a separating set within that number of steps.

### 4.3 Search strategy for $D$-sep links

There is just one more aspect before we can turn the previous results into a complete causal discovery algorithm: Lemma 7 assumes that a minimal separating set is known for each separable pair in the ancestors of the (possible) $D$-sep link $X \leftrightarrow Y$. If these in turn also contain $D$-sep links, as for example $X \leftrightarrow Z$ in Figure 5(b), then we may need to find one before we can find the other. Therefore, if we find a new $D$-separating set we have to check previously tried possible $D$-sep edges in case the new set introduces more nodes into the corresponding hierarchies, after updating $\mathcal{G}^+$. By Lemma 5, as long as there are undiscovered $D$-sep links (still corresponding to edges in $\mathcal{G}^+$), then there is always at least one that cannot have undiscovered $D$-sep links between its ancestors, and so can be recognized as the pattern in Lemma 4 in the (updated) augmented skeleton $\mathcal{G}^+$. As a result, the procedure is guaranteed to terminate only after all are found. Revisiting possible $D$-sep links incurs an additional complexity factor $N^2$.

Finally, in the discussions so far we largely ignored the impact of possible selection bias, as all proofs remain valid with or without selection effects, see details in (Claassen et al., 2013). From (Richardson and Spirtes, 2002, §4.2.1) we know that selection on a node effectively destroys arrowheads on its ancestors, i.e. removes detectable non-ancestor relations. In particular, adding selection effects can overrule the $D$-sep pattern from Lemma 4, and either turn it into a regular separable pair (found by PC) or destroy the independence altogether. As a result, $D$-sep links are much rarer when selection bias is present, and the search for $D$-sep nodes can often be avoided altogether.

## 5 Implementing C-LEARN as FCI+

We incorporate the previous results into a modified **D-SEP** search of the FCI algorithm to obtain a sound and complete method for constraint-based causal discovery that is worst-case polynomial in the number of independence tests between $N$ variables, provided the model is sparse (bounded by $k$). It assumes faithfulness and an underlying causal DAG, but does allow for latent variables and selection bias.

---
**Algorithm 2** FCI+ algorithm
---
    **In** : variables **V**, oracle $\mathcal{O}$, sparsity $k$
    **Out**: causal model $\mathcal{G}$ over **V**
1:  $\mathcal{G}, \mathcal{I} \leftarrow PCAdjSearch(\mathbf{V}, \mathcal{O}, k)$
2:  $\mathcal{G}^+ \leftarrow AugmentGraph(\mathcal{G}, \mathcal{I}, \mathcal{O})$
3:  $PosDsepLinks \leftarrow GetPDseps(\mathcal{G}^+)$
    $D$-SEP search
4:  **while** $PosDsepLinks \neq \varnothing$ **do**
5:     $X, Y \leftarrow Pop(PosDsepLinks)$
6:     $BaseX \leftarrow Adj(X)_{\setminus Y}$
7:     $BaseY \leftarrow Adj(Y)_{\setminus X}$
8:     **for** $n = 1..k$ **do**
9:       **for** $m = 1..k$ **do**
10:         get subset $\mathbf{Z}_X \subseteq BaseX$, size $n$
11:         get subset $\mathbf{Z}_Y \subseteq BaseY$, size $m$
12:         $\mathbf{Z}^* \leftarrow HIE(\{X,Y\} \cup \mathbf{Z}_X \cup \mathbf{Z}_Y, \mathcal{I})_{\setminus \{X,Y\}}$
13:         **if** $X \perp\!\!\!\perp Y \,|\, \mathbf{Z}^*$ **then**
14:           $\mathbf{Z} \leftarrow MinimalDsep(X, Y, \mathbf{Z}^*)$
15:           $\mathcal{I} \leftarrow UpdateSepsets(\mathcal{I}, X, Y, \mathbf{Z})$
16:           $\mathcal{G}^+ \leftarrow AugmentGraph(\mathcal{G}^+, \mathcal{I}, \mathcal{O})$
17:           $PosDsepLinks \leftarrow GetPDseps(\mathcal{G}^+)$
18:           (continue **while**)
19:         **end if**
20:       **end for**
21:     **end for**
22: **end while**
23: $\mathcal{G} \leftarrow RunOrientationRulesFCI(\mathcal{G}, \mathcal{I})$
24: **return** causal model $\mathcal{G}$

---

The FCI+ algorithm in Algorithm 2 starts in line 1 from the output of the PC adjacency search (line 16 in Algorithm 1), that is the skeleton $\mathcal{G}$ and minimal *Sepset* in $\mathcal{I}$ for each eliminated edge. Line 2 constructs the subsequent *augmented skeleton* $\mathcal{G}^+$ by testing for single node additions that destroy the independence, which is the basis for identifying the edges corresponding to possible $D$-sep links in line 3. This list is processed (and updated along the way) until no more unchecked possible $D$-sep links remain. For a pair of nodes $X \leftrightarrow Y$ on a possible $D$-sep edge in $\mathcal{G}^+$ the '*Base*' of adjacent nodes (possible ancestors) is determined in lines 6/7. For each combination of max. $k$ nodes from this base around $X$ and max. $k$ nodes from the base around $Y$ the corresponding hierarchy is computed, and tested for independence in line 13. If found it is turned into a minimal separating set, and stored in the list of sepsets $\mathcal{I}$. This is used to update the augmented skeleton $\mathcal{G}^+$ (remove edge and check for single node dependencies), and to update the set of possible $D$-sep links, e.g. in case we have to reconsider previously rejected edges or can now eliminate other candidates. Finally line 23 runs the standard FCI-orientation rules on the skeleton to return the target causal model in the form of a complete PAG.

**Complexity analysis**

Now to derive a bound on the worst-case complexity of the FCI+ algorithm. As stated, we assume a causal model in the form of a completed PAG $\mathcal{G}$ over $N$ (observed) variables, with node degree $\leq k$, and a constant-time independence oracle.

Contributions of various stages in Algorithm 2:

- l.1: *PCAdjSearch* is order $O(N^{k+2})$, as it searches for subsets $\leq k$ nodes from $N-2$ variables to separate $N^2$ nodes on edges (Spirtes et al., 2000)
- l.2: *AugmentGraph* is order $O(N^3)$, given tests for at most $N-2$ nodes for $1/2N^2$ eliminated edges,
- l.3: *GetPDseps* could find up to $O(N^2)$ possible edges to process,
- l.10-12: there are $O(N^{2k})$ different (implied) combinations for possible $D$-sep sets in the hierarchy from two subsets of at most $k$ nodes from $N-2$ variables,
- l.14: *MinimalDsep* is at most order $O(N^2)$, as nodes can be removed one-by-one (Tian et al., 1998)
- l.16: *AugmentGraph* is again order $O(N^3)$,
- l.17: *GetPDseps* might put back previously tried-but-failed $D$-sep edges, leading to a factor $O(N^2)$,
- l.23: *RunOrientationRulesFCI* is order $O(N^4)$: for $N^2$ edge marks it checks for certain paths in $\mathcal{G}$ which can be done in $O(N^2k)$ by a 'reachability' algorithm, with $Nk \sim$ nr. of edges, (Zhang, 2008, p.1881)

The dominant terms are $O(N^{2k})$ possible hierarchies to test for each of $O(N^2)$ possible edges which may have to be repeated $O(N^2)$ times, leading to an overal worst-case complexity of order $O(N^{2(k+2)})$. In other words: worst-case $O(FCI+) \sim O(PC^2)$, but both with a running time that is polynomial in the number of variables $N$ given maximum node degree $k$.

Clearly, these bounds can be tightened, and many steps can be implemented more efficiently. For example the *Augmentgraph* procedure can be realized using rules on the available $\mathcal{I}$ (instead of additional independence tests), and include invariant tails as well. The number of $BaseX/Y$ combinations to test can be reduced by eliminating adjacent nodes that cannot be ancestor of $X$ or $Y$, and making sure to always include necessary nodes. The hierarchy can be pre-computed for each pair of nodes, and only needs a (partial) update when a new $D$-sep link is found, etc. However, we are not (yet) looking for an optimal implementation, just to verify that a polynomial solution is possible.

## 6  Discussion

We have shown that it is possible to learn a sound and complete sparse causal model representation that is polynomial in the number of independence tests, even when latent variables and selection bias may be present. We presented an implementation in the from of the FCI+ algorihtm, which derives from standard FCI but with a modified *D-SEP* search stage.

The exponential complexity can be avoided by exploiting the inherent structure in the problem that stems from the underlying causal network. The problem can be broken down into a series of smaller steps, in contrast to, e.g. finding a minimal Bayesian network or a travelling salesman problem, where the minimal solution only applies to the specific network as a whole. This 'breaking down into subproblems' follows naturally in the constraint-based paradigm; an intriguing question is whether it is also possible to find a polynomial score-based causal discovery method.

Effectively, FCI+ uses independence information from previous stages to guide the *D-SEP* search, leading to a theoretical worst-case running time $O(N^{2(k+2)})$ in the number of independence tests. This is significantly better than exponential, but still unfeasible for large $N$. However, in practice the typical performance is vastly better: very few candidate $D$-sep links with even fewer repeats, and in particular much smaller sizes for the adjacent set $An(\{X,Y\}) \sim O(2k)$ for $BaseX/Y$ (line 6/7 in Algorithm 2). This leads to a dramatic reduction in actual runtime, as the most expensive term reduces from $O(N^{2k})$ to $O(2^{2k})$ base combinations for the $D$-sep set candidates.

A drawback for sample versions of FCI+ is that the conditioning sets may become larger than strictly necessary, leading to a loss of statistical power of the independence tests. Preliminary investigations in random graphs up to 200 variables suggest this effect is not very prominent (often the largest set remains smaller than for standard FCI), but it may become an issue in certain circumstances. An interesting solution is to limit the maximum node sets by taking the *intersection* of the hierarchical **D-SEP** candidate and FCI's **Possible-D-SEP** sets. Furthermore, many tests in the AugmentGraph procedures (line 2,16) may be avoided by applying some of FCI's orientation rules to the available structure. Finally we can try to restrict the number of possible $D$-sep-edges to check even further, and/or optimize the sequence in which to process them. Ultimately our goal is to come as close to the PC runtime as possible, while still handling hidden variables correctly.

### Acknowledgements

Tom Claassen and Tom Heskes were supported by NWO Vici grant nr.639.023.604; Joris Mooij was supported by NWO Veni grant 639.031.036.

## Appendix A. Proofs

This section contains proof sketches for the results in this paper; for details and examples, see (Claassen et al., 2013).

**Lemma 3.** *(D-separated nodes are non-ancestors)*

*Proof sketch.* In the supplement we show that if $X$ is not independent of $Y$ for any subset $Adj(X)$, then $Y$ is not an ancestor of $X$ and has no selection bias. For a $D$-sep link $(X,Y)$ this applies to both; statement for minimal $\mathbf{Z}$ follows immediately from Lemma 2-(2). □

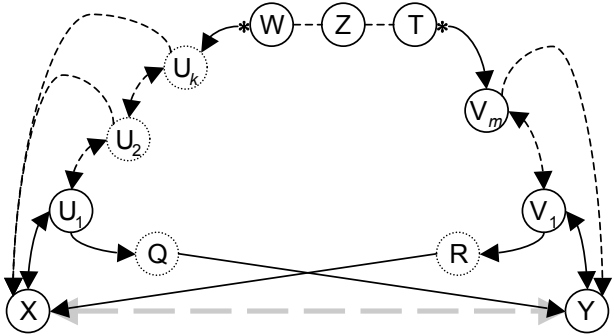

Figure 7: Path configuration for $D$-sep link $X \cdots Y$.

**Lemma 4.** *(bi-directed D-sep patterns in $\mathcal{G}^+$)*

*Proof sketch.* In the supplement we show that the path configuration in Figure 7 is always present for a $D$-sep edge $X - Y$, resulting in identifiable independencies $X \perp\!\!\!\perp W \mid [\mathbf{Z}']$, $Y \perp\!\!\!\perp T \mid [\mathbf{Z}'']$, and $U_1 \perp\!\!\!\perp V_1 \mid [\mathbf{Z}''']$ (for some sets $\mathbf{Z}'$ etc.). Furthermore, the first necessarily becomes dependent when including either $U_1$ or $Y$ into the separating set. Idem for the second when including $V_1$ or $X$, as for the third when including $X$ or $Y$. In combination with Lemma 2-(1) this implies identifiable invariant arrowheads in $\mathcal{G}^+$ on all three edges $U_1 \leftrightarrow X \leftrightarrow Y \leftrightarrow V_1$, with $U_1$ not adjacent to $V_1$. □

**Lemma 5.** *(D-sep link hierarchy)*

*Proof sketch.* Follows from Lemma 3, otherwise it would introduce a cycle. □

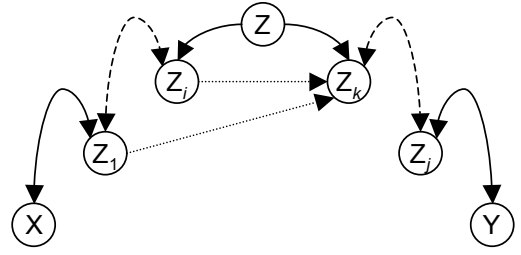

Figure 8: Canonical path blocked by $D$-sep node $Z$.

**Lemma 6.** *(all required D-sep nodes appear in other minimal separating sets we already found)*

*Proof sketch.* In the supplement we show that each $D$-sep node $Z$ blocks a path of the generic form depicted in Figure 8. From this we construct a necessary non-adjacency of neighboring node $Z_k$ with at least one of the other $\{Z_1,..,Z_i\}$ or $X$ along the path, in which $Z$ is part of a minimal set that separates them. □

**Lemma 7.** *(for D-sep link $X \leftrightarrow Y$ in $\mathcal{G}^+$ and independence set $\mathcal{I}$, the set $HIE(AA(\{X,Y\}),\mathcal{I})_{\setminus\{X,Y\}}$ is a D-separating set)*

*Proof sketch.* In the supplement we show that for a $D$-sep link the set of nodes in the hierarchy $HIE(\{X,Y\},\mathcal{I})$ is independent of the specific (minimal) independence found for each pair of nodes that are not adjacent in $\mathcal{G}^+$. We do this by showing that if a node $W$ is *optional* in a minimal separating set between two nodes $X$ and $Y$, so both $X \perp\!\!\!\perp Y \mid [\mathbf{Z}]$ and $X \perp\!\!\!\perp Y \mid [\mathbf{W}]$ exist, with $W \in (\mathbf{W} \setminus \mathbf{Z})$, then there is also an optional node $Z \in (\mathbf{Z} \setminus \mathbf{W})$, and $W$ is part of a minimal separating set between $Z$ and either $X$ or $Y$. This argument can be repeated until the node is necessary in some minimal separating set, and will be found in all independence sets $\mathcal{I}$. So every node that is optional in one minimal separating set (and so may not be included in the corresponding Sepset from the PC adjacency search) is a necessary node in a Sepset between some other pair when 'zooming in' on the graph. As these are all included, it follows that $HIE(\{X,Y\},\mathcal{I})$ is independent of the specific separating sets found in $\mathcal{I}$. By Lemma 6, all $D$-sep nodes we need to separate $X$ and $Y$ als appear in some minimal set between two nodes from the union of $\{X,Y\}$ and the corresponding $D$-separating set, for which the same applies again, until ultimately an independence between two nodes in $Adj(\{X,Y\})\cup\{X,Y\}$ is reached. But that means they are all part of $HIE(AA(\{X,Y\}),\mathcal{I})_{\setminus\{X,Y\}}$, and so this is guaranteed to be a $D$-separating set for $X \leftrightarrow Y$. □